\documentclass[12pt]{iopart}

\usepackage{graphicx}
\usepackage{cite}
\usepackage{amsfonts} 
\usepackage{xcolor}

\begin{document}

\title[Dynamical loss functions shape landscape and improve learning in ANNs]{Dynamical loss functions shape landscape topography and improve learning in artificial neural networks}

\author{Eduardo Lavín-Pallero$^1$ and Miguel Ruíz-García$^{2,3}$}
\address{$^1$Departamento de Matemática Aplicada, ETSII, Universidad Politécnica de Madrid, Madrid, Spain}
\address{$^2$Departamento de Estructura de la Materia, Física Térmica y Electrónica, Universidad Complutense Madrid, 28040 Madrid, Spain}
\address{$^3$GISC - Grupo Interdisciplinar de Sistemas Complejos, Universidad Complutense Madrid, 28040 Madrid, Spain}
\ead{elavinpal@gmail.com, miguel.ruiz.garcia@ucm.es}
\vspace{10pt}
\begin{indented}
\item[] May 2025
\end{indented}

\begin{abstract}
Dynamical loss functions are derived from standard loss functions used in supervised classification tasks, but are modified so that the contribution from each class periodically increases and decreases. These oscillations globally alter the loss landscape without affecting the global minima. In this paper, we demonstrate how to transform cross-entropy and mean squared error into dynamical loss functions. We begin by discussing the impact of increasing the size of the neural network or the learning rate on the depth and sharpness of the minima that the system explores. Building on this intuition, we propose several versions of dynamical loss functions and use a simple classification problem where we can show how they significantly improve validation accuracy for networks of varying sizes. Finally, we explore how the landscape of these dynamical loss functions evolves during training, highlighting the emergence of instabilities that may be linked to edge-of-instability minimization.
\end{abstract}

%
%
%
%
%

\section{Introduction}
Most machine learning tasks can be reduced to the search of extrema in a loss function in a high-dimensional parameter space~\cite{rosasco2004loss,choromanska2015loss,li2018visualizing,soudry2016no,cooper2018loss,verpoort2020archetypal,janocha2017loss, kornblith2020s,ballard2017energy,mannelli2019afraid,arous2019landscape,goldblum2019truth,niroomand2024explainable,niroomand2024insights}. Although this picture may look reminiscent of several problems in physics, e.g. the minimization of the energy of a system of soft-interacting particles or the optimization of physical networks~\cite{franz2016simplest,geiger2019jamming,franz2019jamming,franz2019critical,geiger2020scaling,geiger2020perspective,ruiz2017bifurcation,rocks2017designing,ruiz2019tuning,yan2017architecture,ruiz2021emergent,martinez2024fluidic,altman2025collective,ruiz2020topologically}, the optimization (training) of neural networks (NNs) challenges our intuition. Contrary to the smooth minimization of physical properties that we may have in mind, \textcolor{black}{NN} minimization is characterized by instabilities. In particular, machine learning minimization typically occurs in regions of the landscape where only a few eigenvalues of the Hessian are clearly positive, whereas the rest of the eigenvalues remain close to zero or even take negative values \cite{sagun2017empirical,sagun2016eigenvalues,sankar2021deeper,yao2020pyhessian,ahn2022understanding,kaur2023maximum}. Within the subspace spanned by these outliers of the Hessian, the loss function landscape typically displays a valley-like structure, and the model descends it towards regions of smaller values of the loss. During this minimization process, the model oscillates between the walls of the valley~\cite{xing2018walk} and can become trapped if it reaches a region where the valley is too narrow for the minimization step to proceed. In such cases, if the model fails to find a wider valley, it can lead to \textit{edge of stability} minimization~\cite{cohen2022gradient, zhu2022understanding, arora2022understanding}.

The ubiquity of edge-of-stability minimization in deep learning~\cite{cohen2022gradient}, at odds with the intuition gained from most physical systems, presents an opportunity for new understanding and for the exploration of alternative optimization techniques. 
We think that dynamical loss functions~\cite{ruiz2021tilting} can allow us to tackle both questions.
These loss functions -- weighted by class with weights that oscillate during minimization -- have been shown to improve training at the same time that they transform the topology of the landscape~\cite{ruiz2021tilting}; and they 
are reminiscent of other oscillating strategies in biological systems, materials science and physics that lead to better solutions~\cite{murugan2019bioinspired, murugan2021roadmap, mustonen2009fitness, di2024emergent, cairns2022strong, falk2023learning}.

In deep learning, there are other methods that change the loss function during learning, a paradigmatic example is curriculum learning~\cite{bengio2009curriculum}, where the training examples are not randomly presented but are organized in a meaningful order. In practice, this ``curriculum'' can be achieved by weighing the contribution of easier samples (e.g. most common words) to the loss function more at the beginning and increasing the weight of more difficult samples (e.g. less frequent words) at the end of training. In this way one expects to start with a smoothed-out version of the loss landscape that progressively becomes more complex as training progresses. Curriculum learning has played a crucial role across deep learning~\cite{pmlr-v48-amodei16, graves2016hybrid,silver2017mastering,lozano2025towards,strittmatter2025curriculum}. However, its main drawback is that it requires additional supervision: a second label for the difficulty or complexity of the data. Even more, if continuously changing the landscape facilitates learning, why do it only once? Furthermore, training data is already divided into different classes---is it possible to take advantage of this already-existing label for each training example instead of introducing a new label for difficulty? These questions motivated the creation of dynamical loss functions~\cite{ruiz2021tilting}. We think that studying the interplay between edge-of-stability minimization and dynamical loss functions can shed light on how the topography of the loss function landscape shapes learning and how modifying this topography can improve the behavior of deep learning.

\section{The role of landscape topography}

In supervised classification tasks, the loss function $\mathcal{F}$ is defined such that its global minima ($\mathcal{F}=0$) correspond to the correct classification of the \textit{training} dataset. In the case of cross-entropy:

\begin{equation}
    \mathcal{F}_{\mathrm{CE}}= \sum_{j \leq P}    - \log \left(\frac{e^{f_{y_j}(x_j,\mathbf{W})}}{\sum_{i}e^{f_i(x_j,\mathbf{W})}} \right),
    \label{eq_loss}
\end{equation}

where $x_j$ is an element of the training set of size $P$, $y_{j}$ is its label/class, and $f(x_j,\mathbf{W})\in\mathbb{R^C}$ is the \textcolor{black}{function representing the neural network, where we explicitly indicate the inputs: $x_j$, and $\mathbf{W}$ (the trainable parameters); finally, $C$ is the number of classes.}
Learning in this type of problems boils down to the minimization of the loss function, and reaching the global minima strongly depends on the number of parameters present in the neural network ($\mathbf{W}$) compared with the amount (or complexity) of the data. Classic bias-variance tradeoff suggests that too few parameters ($\mathbf{W}$) will not be enough to learn the data (underparametrized regime), whereas too many parameters will lead to overfitting, damaging generalization (overparametrized regime). However, neural networks have been shown to generalize best in the overparametrized regime~\cite{belkin2019reconciling,zhang2021understanding,nakkiran2021deep,d2020double,rocks2022memorizing}. This paper shows how dynamical loss functions reduce the critical number of parameters that leads to overparameterization in a simple classification problem, allowing the reduction of computational costs and potentially improving generalization. 

\begin{figure}
    \centering
    \includegraphics[width=\linewidth]{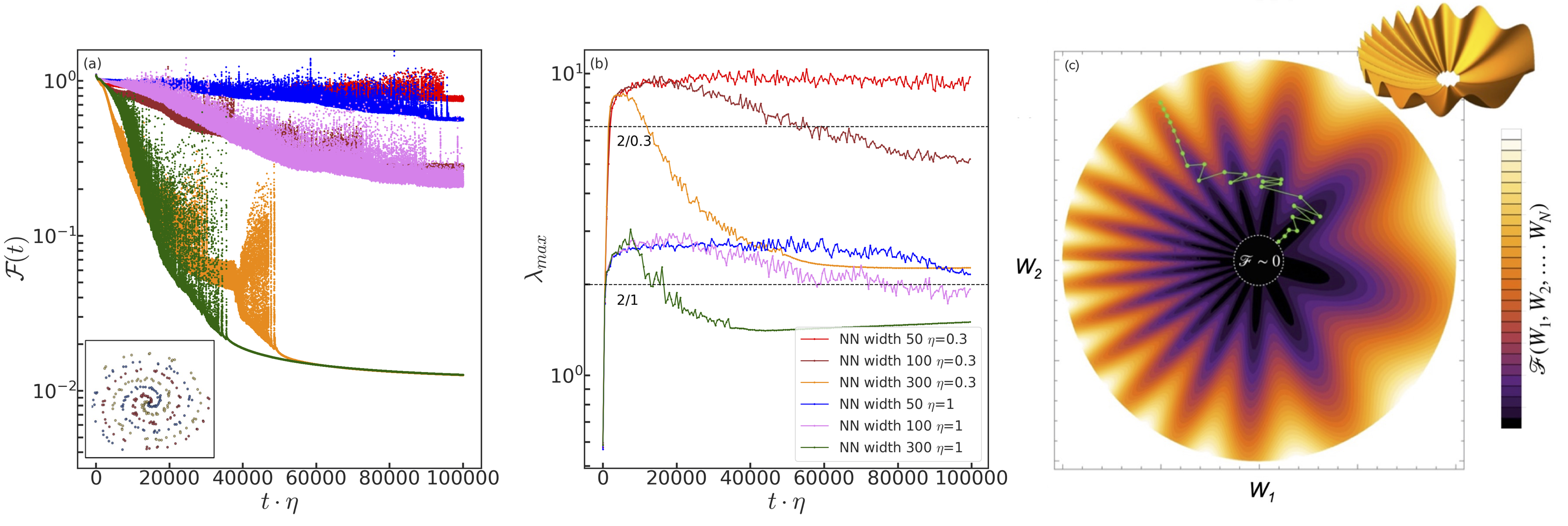}
    \caption{
    Interplay between network size and learning rate in the minimization of cross-entropy loss function. For three different network sizes and two different learning rates ($\eta$) we average the instantaneous values of the loss function and curvature (largest eigenvalue of the Hessian, $\lambda_{max}$) during the minimization of 20 simulations. The training dataset is presented in the inset of panel (a)---2D points belonging to three classes that follow the three arms of a spiral. Panel (a) shows the mean value of the loss, whereas (b) shows the mean value of the largest eigenvalue of the Hessian. Spikes during training (edge-of-stability minimization) are more clear in panel (a). Although panel (a) suggests that the value of the loss during training is controlled by NN size, panel (b) shows how smaller learning rates tend to explore narrower regions of the landscape. Even so, we see that for large network sizes the system tends to reach wider valleys for both learning rates, scaping edge-of-stability minimization. Panel (c) shows an idealized illustration of the loss function landscape that can qualitatively explain the behavior of panels (a) and (b). We represent a subspace of parameter space, that displays in its center a global minimum that occupies a finite region of parameter space. Valleys that lead to the global minimum need to increase their sharpness to fit into the boundary. Minimization in this landscape (green dots) can lead to instabilities due to the increase in curvature as the system approaches the minimum. During the instabilities the system can jump towards wider valleys and in some cases it can find one that is wide enough to smoothly descend into the global minimum (this would occur if the NN is large enough). Panel (c) is only a toy idealization to help gain intuition about learning in \textcolor{black}{NNs}. }
    \label{fig_1}
\end{figure}

First, we study how minimization develops for different network sizes and learning rates in the Swiss Roll classification problem, aiming to gain valuable intuition that will help us later to better understand the learning process with dynamical loss functions. The task is to classify points in 2D from a three-armed spiral according to their colors (see inset of Fig.~\ref{fig_1} (a)), and we use standard (non-dynamical) cross-entropy. We use a fully connected neural network (NN) with one hidden layer of variable width. We compute the Hessian of the loss during minimization, where its largest eigenvalue indicates the \textit{sharpness}~\cite{gur2018gradient,cohen2022gradient}. Fig. \ref{fig_1} displays the results obtained for three NN widths and two different learning rates. Panel (a) shows how learning improves as the network size increases---larger network sizes reach deeper and wider basins where the minimization process stabilizes. However, the results also highlight the improvement in learning as the learning rate increases. Panel (b) shows how the local curvature evolves during training. For all the cases, minimization starts in regions with smooth curvature and the system quickly moves to a region where the largest eigenvalue is slightly above $2/\eta$. This phenomenon is consistent with edge-of-stability (EoS) minimization. For the largest network, the system finds a way to escape EoS and enters a region of high curvature leading to better results. Panel (c) provides a qualitative explanation of this phenomenon with the only purpose of gaining valuable intuition to understand the minimization with dynamical loss functions in the next sections. We hypothesize that typical loss function landscapes in NNs are characterized by global minima that occupy bounded regions of parameter space and are surrounded by valleys that descend towards them (this is consistent with observations in recent work~\cite{annesi2023star}). Due to the finite size of the $\mathcal{F} \sim 0$ regions, the valleys need to increase their sharpness to fit along the boundary of the global minima. As the system descends a valley (green dots) the curvature increases until the minimization becomes unstable. Then, the NN can jump sequentially from one valley to another, descending the loss landscape while maintaining approximately the same curvature. Increasing the NN width leads to a parameter space of higher dimension and wider valleys. When the width is high enough, the NN will explore several valleys (EoS phase) until it finds a valley that is wide enough to descend to the global minimum. Even more, larger learning rates explore larger portions of the landscape as they try to find a valley that is wide enough, which they descend more quickly. Counterintuitively, minimization instabilities can be harnessed to find deeper regions of the landscape~\cite{lewkowycz2020large,cohen2022gradient,ruiz2021tilting}.

\section{Dynamical loss functions}

We show now how dynamical loss functions~\cite{ruiz2021tilting} can lead to sequential instabilities and a larger exploration of parameter space, improving generalization and allowing us to use smaller NNs. First we carry out a simple variation of cross-entropy, that we will denote dynamical cross entropy (DCE):

\begin{equation}
    \mathcal{F}_{\mathrm{DCE}}= \sum_{j \leq P} \Gamma_{y_j}(t)  \left( - \log \left(\frac{e^{f_{y_j}(x_j,\mathbf{W})}}{\sum_{i}e^{f_i(x_j,\mathbf{W})}} \right) \right)
    \label{eq_loss}
\end{equation}

Where $\Gamma_i$ is a different oscillating factor for \textit{each class $i$}, and $t$ is the number of minimization steps (see panel (a) in Fig.~\ref{fig_2} for one example).  The specific expression for $\Gamma_i(t)$ are detailed in the Supplementary Materials.  Depending on the values of $\Gamma_i$, the topography of the loss function will change, but the loss function will still vanish at the same global minima, which are unaffected by the values of $\Gamma_i$. 
We use $\Gamma_i$ to emphasize one class relative to the others for a period $T$, and cycle through all the classes in turn so that the total duration of a cycle that passes through all classes is $CT$.  Figure \ref{fig_2} (a) shows the oscillating factors $\Gamma_i$ that we use in the case of DCE for the case of the dataset with three classes shown in the inset of Fig. \ref{fig_1} (a), these oscillations are controlled by two parameters, their amplitude ($A$) and period ($T$). In the case $A \gg 1$ the network will output the chosen class regardless of input. However, in the next period the network will have to learn a different class, suggesting that the transition between periods will mark the points at which the topography of the landscape becomes more complex---note that right at the transition all $\Gamma_i$ are $1$ and we recover the standard loss function. \textcolor{black}{ The Supplementary Materials include a phase diagram showing how $A$ and $T$ lead to different final accuracies.}

\begin{figure*}
    \centering    
    \includegraphics[width=\linewidth]{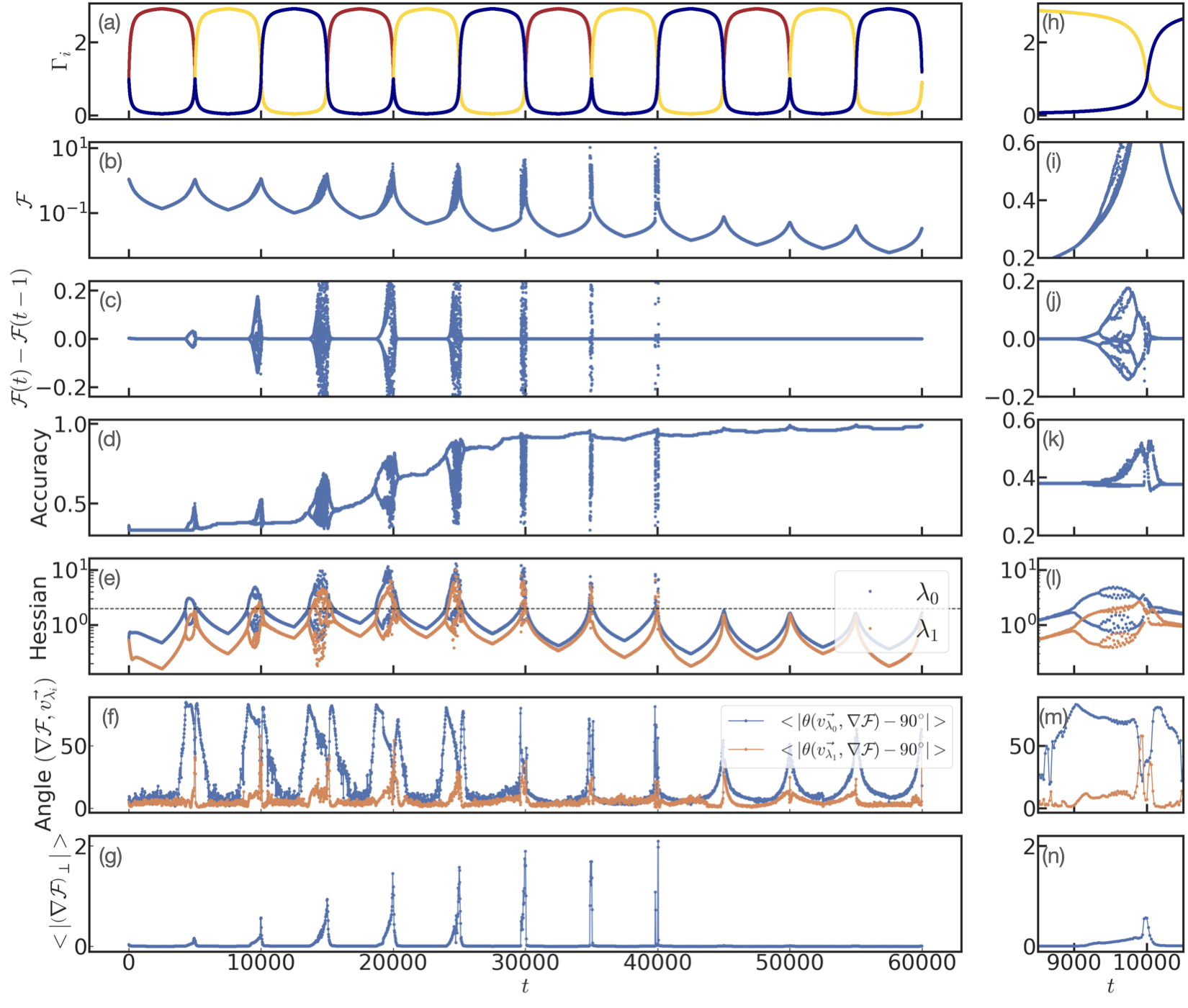}
    \caption{Evolution of learning with a dynamical loss function. Panel (a) shows the evolution of the parameters $\Gamma_i$ which control the change of the dynamical loss function $\mathcal{F}(t)$: $\Gamma_1$, $\Gamma_2$ and $\Gamma_3$ correspond to red, yellow and blue solid lines.  Panel (b) displays the instantaneous value of the loss during training, panel (c) shows the local change of $\mathcal{F}(t)$, removing the general trend and emphasizing the instabilities, whereas panels (d) and (e) show the training accuracy and the two largest eigenvalues of the Hessian ($\lambda_1$ and $\lambda_2$), respectively. \textcolor{black}{Panel (f) represents the angle formed by the gradient and the two eigenvectors of the Hessian ($v_{\lambda_1}$ and $v_{\lambda_2}$), we compute the module of the difference with $90^\circ$ and average over 10 time steps. Panel (g) represents the module of the projection of the gradient on the subspace perpendicular to $v_{\lambda_1}$ and $v_{\lambda_2}$, also averaged for 10 time steps. 
    Panels (h)-(n)} present a zoom of one of the period doubling cascades. We use a neural network with one hidden layer of width $100$ and full batch gradient descent, where $T=5000$ and $A=70$, chosen for ease of visualization.}
    \label{fig_2}
\end{figure*}

\begin{figure*}
    \centering    
    \includegraphics[width=\linewidth]{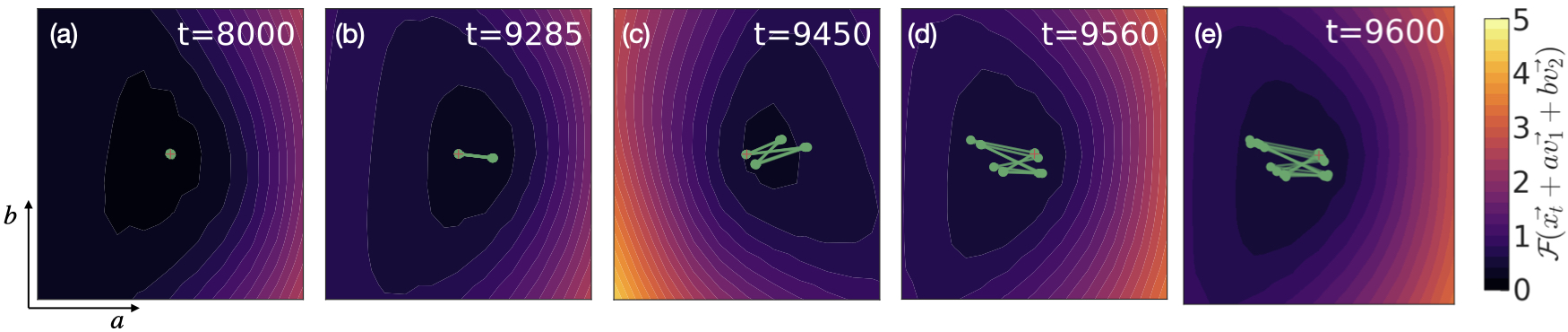}
    \caption{\textcolor{black}{Visualizing the evolution of the system during one period doubling cascade. Panels (a)-(e) display the evolution of the system as it approaches the last part of the second period in Fig. \ref{fig_2} ($t=10000$), where instabilities emerge. We represent a subspace of the parameter space spanned by the eigenvectors associated with the two largest eigenvalues of the Hessian, $\vec{v}_1$ and  $\vec{v}_2$, where $a$ and $b$ take values in $[-1,1]$. The background color displays the value of $\mathcal{F}$ around the point in parameter space where the system is at $t=8000$, $9285$, $9450$, $9560$ and $9600$ ($\vec{x}_t$, a red cross in the plots). We fix the background and plot green dots representing the position of the system at times $\{t, t+1, \dots, t+20 \}$, connecting them with semitransparent green arrows. We use a neural network with one hidden layer of width $100$ and full batch gradient descent, where $T=5000$ and $A=70$, chosen for ease of visualization.}}
    \label{fig_2_bis}
\end{figure*}

\section{Results and discussion}

The behavior of the system as it locally minimizes the dynamical loss function landscape is summarized in Fig.~\ref{fig_2}. In the first half of each period $T$ the instantaneous value of $\mathcal{F}_{\textrm{DCE}}$ and the largest eigenvalues of the Hessian (curvature) decrease, whereas the accuracy stays flat. The system is descending a valley that shifts downwards and also widens. 
In the second half of each oscillation, the valley narrows and rises ($\mathcal{F}$ increases even though the system is undergoing gradient descent; see panel (b)). Additionally,  something remarkable happens when the largest eigenvalue of the Hessian crosses the value $2/\eta$, marked by the horizontal dashed line in panel (e): a bifurcation emerges, clearly visible in panels (c) and (d). There are additional bifurcations each time another eigenvalue reaches $2/\eta$, forming a period-doubling cascade, \textcolor{black}{see panels (j)-(i) for a zoom of one of this cascades.} These bifurcations correspond to the system bouncing between the walls of the valley, this is clearly depicted by \textcolor{black}{Fig. \ref{fig_2_bis}, which} shows a section of parameter space along the eigenvectors associated to the two largest eigenvalues of the Hessian. \textcolor{black}{Fig. \ref{fig_2_bis} (a) (corresponding to $t=8000$) shows how the state of the system (represented by green dots) stay in the same position in this section of parameter space for 20 time steps. This suggests that in the regime of stable minimization (before unstabilities emerge), the system moves along the bottom of a valley where the gradient is perpendicular to the eigenvectors associated to the two largest eigenvalues of the Hessian, in agreement with the toy representation shown in Fig. 1 c. As time approaches $t=10000$ (Fig \ref{fig_2_bis} (b)-(e)), the curvature increases and the minimization becomes unstable,} and we see how the system starts bouncing between two walls, four, etc\dots Remarkably, these instabilities do not destroy learning but seem to enhance it
---note how \textcolor{black}{accuracy increases during the first few bifurcation cascades, although this effect vanishes once the system reaches high accuracy}. \textcolor{black}{We quantify this effect in panels (f) and (g) of Fig.~\ref{fig_2}. Panel (f) shows that indeed, during smooth minimization, the gradient is perpendicular to the subspace spanned by \(v_{\lambda_1}\) and \(v_{\lambda_2}\)---the two eigenvectors associated with the largest eigenvalues of the Hessian. In contrast, when oscillations appear, the gradient initially aligns with \(v_{\lambda_1}\) and later with \(v_{\lambda_2}\). More importantly, panel (g) demonstrates that although such alignments do occur, the gradient magnitude can also increase substantially. Notably, while the gradient aligns with these two directions during the instabilities, panel (g) shows that the projection of the gradient onto the subspace perpendicular to \(v_{\lambda_1}\) and \(v_{\lambda_2}\) also grows markedly toward the end of the oscillations---reaching values higher than those observed during smooth minimization. This indicates that the instabilities enhance the exploration in directions perpendicular to those of maximum curvature, reminiscent of the toy scenario illustrated in Fig. \ref{fig_1}(c).}

\begin{figure}
    \centering
    \includegraphics[width=1\linewidth]{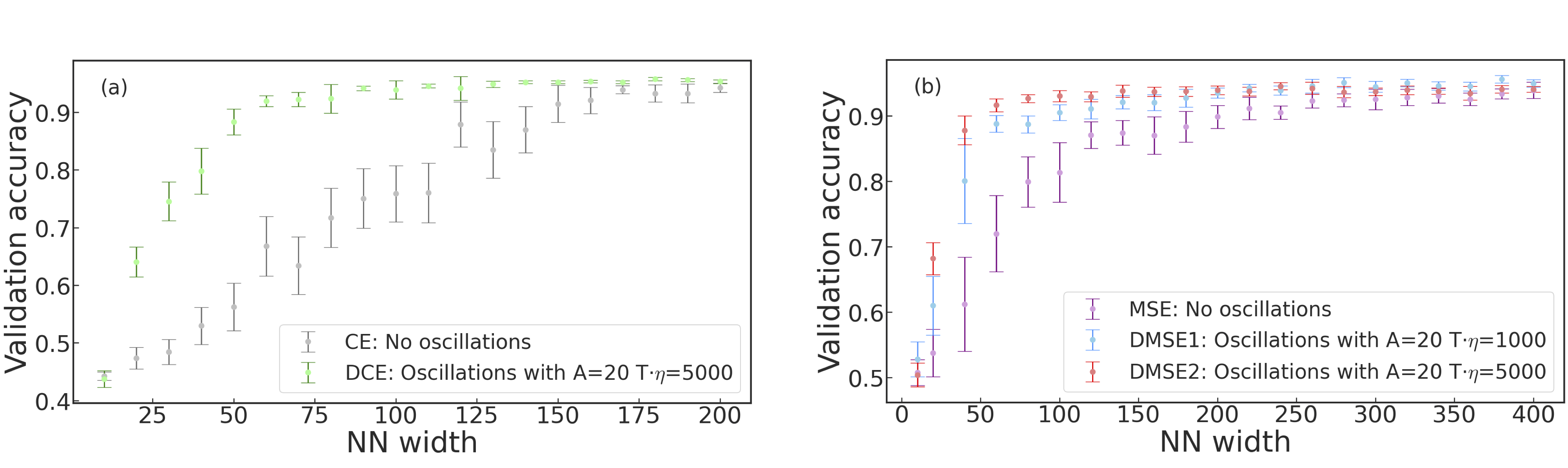}
    \caption{Dynamical loss functions can train smaller networks. We use again a fully connected \textcolor{black}{NN} and the Swiss Roll dataset as in Fig. \ref{fig_1} (a). The validation dataset in both cases is similar to the training dataset but generated with a different seed leading to a different distribution of the points along the spiral. In panel (a) we compare CE and DCE with $\eta = 1$, and we average 50 simulations for each NN size. In panel (b) we compare MSE with the two versions of DMSE with $\eta = 0.075$, and average 10 simulations for each NN size. In both panels we change $A=1$ (static loss) for the last oscillation where we measure the accuracy, showing that dynamical loss functions take the system to a different region of parameter space (compared to their static versions) where validation accuracy improves.  The errorbars display the error of the mean. }
    \label{fig_3}
\end{figure}

\begin{figure}
    \centering
    \includegraphics[width=0.55\linewidth]{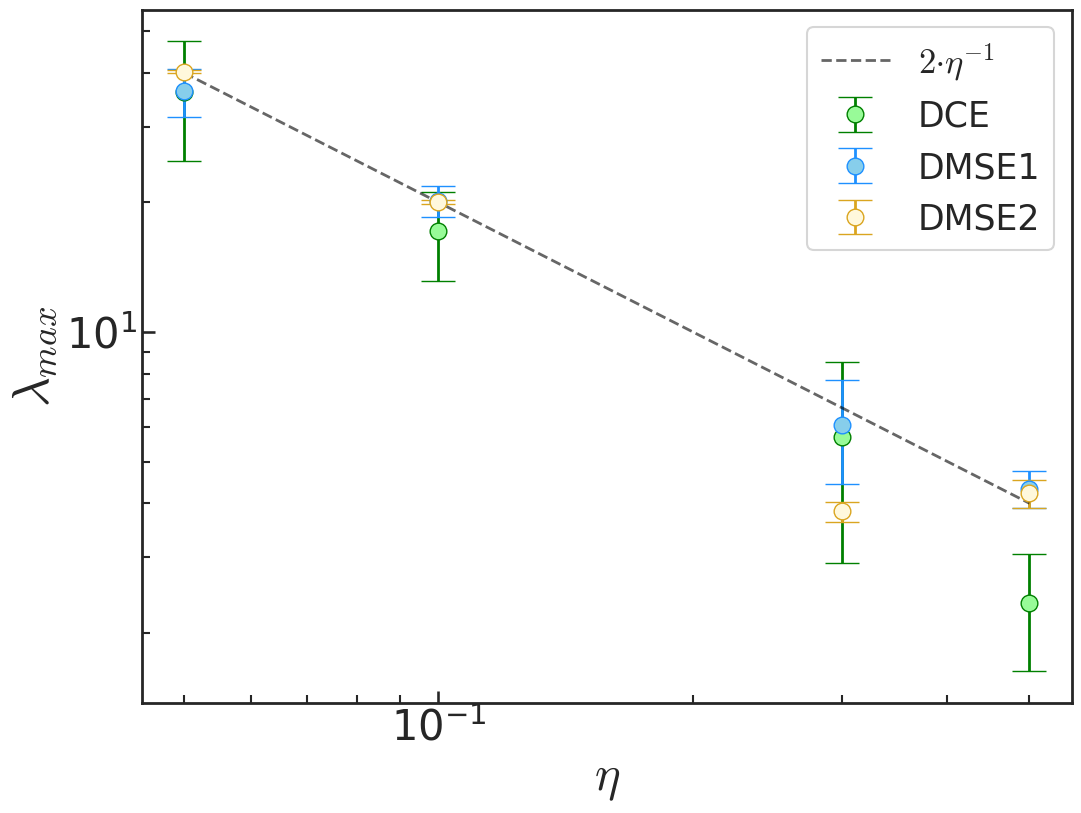}
    \caption{We represent the value of $\lambda_{\mathrm{max}}$ at which instabilities emerge (see Fig. \ref{fig_2} (d)) vs $\eta$. To show the robustness of the result, we simulate different conditions: $A = [25, 50, 75]$, $T = [5{,}000, 10{,}000]$ for a total time $T_{\mathrm{total}} = 100{,}000/\eta$, with  $NN_{\mathrm{width}} = 100$. We represent the mean value of 30 simulations per condition. The dash line is the prediction from edge-of-stability minimization. The error bars display the error of the mean.}
    \label{fig_4}
\end{figure}

We also propose two other dynamical loss functions based on mean square error (MSE):
\begin{equation}
    \mathcal{F}_{\mathrm{DMSE1}} = \frac{1}{N}\sum_{j \leq P} \Gamma_{i}(t) \sum_{i \leq C} \left( f_i(\textit{\textbf{x}}_j,\textit{\textbf{W}}) - \textit{y}_{j,i}  \right)^2,
\label{eq:MSE_1}
\end{equation}
\begin{equation}
    \mathcal{F}_{\mathrm{DMSE2}} = \frac{1}{N}\sum_{j \leq P} \sum_{i \leq C} \left( f_i(\textit{\textbf{x}}_j,\textit{\textbf{W}}) -  \Gamma_{i}(t) \textit{y}_{j,i}  \right)^2,
\label{eq:MSE_2}
\end{equation}
see the Supplementary Materials for a detailed explanation of their behavior. Despite their differences, we observe that DMSE1 and DMSE2 both display bifurcations reminiscent of Fig. \ref{fig_2} and, similarly to DCE, they improve learning and generalization (at least for this simple classification task). Fig. \ref{fig_3} shows how dynamical loss functions improve validation accuracy over their corresponding standard static loss (CE or MSE). This difference is greatest for small network sizes---dynamical loss functions shift the under to over-parameterized transition---but validation accuracy is also higher for the dynamical loss function at large NN sizes. To explore the effect on system size in Fig. \ref{fig_3} we have selected a specific amplitude and period for the oscillations, \textcolor{black}{that correspond to} a region in the hyperparameter space (amplitude and period) that provides better results compared to those obtained with the standard loss functions (see Supplementary Material), although the results can also deteriorate when the amplitude and period become too large, and catastrophic forgetting can occur~\cite{mccloskey1989catastrophic,ruiz2022model}. 
To further explore the similarities between DCE, DMSE1 and DMSE2, and their connection with edge-of-stability minimization, we present in Fig. \ref{fig_4} the relationship between the mean value of the Hessian largest eigenvalue at which the first instability occurs for different learning rates. We see how the relationship between curvature and learning rate displays the same scaling for DCE, DMSE1 and DMSE2, compatible with $2/\eta$, the theoretical prediction from `\textit{edge of stability}' minimization~\cite{cohen2022gradient}.

\section{Conclusion}

This paper explores the behavior of dynamical loss functions, a learning protocol for \textcolor{black}{neural networks} that has deep connections to fundamental problems in physics~\cite{ruiz2019tuning, murugan2019bioinspired, murugan2021roadmap, mustonen2009fitness, di2024emergent, cairns2022strong, falk2023learning, annesi2023star, cohen2022gradient}. We have demonstrated how this approach can enhance the behavior of NNs, reducing the critical size at which overparametrization begins for three different dynamical loss functions in a simple classification problem. Along this paper we have also discussed how the topography of the landscape controls the minimization process, and how dynamical loss functions introduce instabilities that enhance learning. The emergence of these instabilities coincides with the system visiting regions of the landscape that are too narrow for the learning rate, connecting dynamical loss function phenomenology with edge-of-stability minimization. 
\textcolor{black}{We used full-batch gradient descent throughout this work in order to isolate the effects of the dynamical loss function, avoiding the intrinsic stochasticity of stochastic gradient descent (SGD). Nevertheless, it would be interesting to investigate the interplay between dynamical loss functions and SGD, since both approaches enhance exploration in parameter space. In particular, one could design a minibatch selection scheme where the sampling probabilities depend on the class of each example and oscillate in a manner analogous to the $\Gamma_i$ parameters introduced in this paper. Such a protocol could mimic some of the features reported here, while relying solely on the order in which the neural network encounters the data, thereby connecting more explicitly with curriculum learning. Returning to the questions posed in the introduction, we have shown how we can improve learning taking advantage of the classes in which the data is already divided to carry out protocols reminiscent of curriculum learning, but without requiring additional labels or supervision. Moreover, this improvement can be further amplified as the system cycles repeatedly through the same protocol. }

We hope that the dynamical loss functions that we have presented in this work will open up a plethora of possibilities to explore how changing the topography of the loss function landscape during minimization can control (and potentially improve) learning in \textcolor{black}{neural networks}. They allow the practitioner to probe different regions of the landscape that could not be reached with traditional (static) loss functions. We also hope that this work will motivate other researchers to design other dynamical loss functions that could be optimal for other supervised classification tasks. Other dynamical loss functions could be designed in many different ways (we have presented three types here), the simpler way to modify the ones presented here would be to modify the way the $\Gamma_i$ change with time. To understand how different dynamical loss functions affect learning for different tasks is an open question. Finally, we hope that dynamical loss functions can be used by the physics community to understand the connections between learning in artificial neural networks and other minimization protocols in physical systems.

\section{Acknowledgements}

It is a pleasure to acknowledge many stimulating discussions with Ge Zhang, Samuel S Schoenholz and Andrea J Liu and to thank them for their
valuable suggestions.
We acknowledge support from the Ramón y Cajal program (RYC2021-032055-I) funded by MCIN/AEI/10.13039/501100011033 and by European Union NextGenerationEU/PRTR, a Research Grant from HFSP (Ref.-No: RGEC33/2024) with the award DOI (https://doi.org/10.52044/HFSP.RGEC332024.pc.gr.194170) and grant PID2023-147067NB-I00 funded by MCIU/AEI/10.13039/501100011033 and by ERDF/EU. We acknowledge the use of AI-based language tools (ChatGPT, OpenAI) to assist with the editing and refinement of the manuscript's text.

\section{References}

\bibliography{dyn_loss_bib}

\providecommand{\newblock}{}
\begin{thebibliography}{10}
\expandafter\ifx\csname url\endcsname\relax
  \def\url#1{{\tt #1}}\fi
\expandafter\ifx\csname urlprefix\endcsname\relax\def\urlprefix{URL }\fi
\providecommand{\eprint}[2][]{\url{#2}}

\bibitem{rosasco2004loss}
Rosasco L, Vito E~D, Caponnetto A, Piana M and Verri A 2004 {\em Neural
  Computation\/} {\bf 16} 1063--1076

\bibitem{choromanska2015loss}
Choromanska A, Henaff M, Mathieu M, Arous G~B and LeCun Y 2015 The loss
  surfaces of multilayer networks {\em Artificial intelligence and
  statistics\/} (PMLR) pp 192--204

\bibitem{li2018visualizing}
Li H, Xu Z, Taylor G, Studer C and Goldstein T 2018 {\em Advances in neural
  information processing systems\/} {\bf 31}

\bibitem{soudry2016no}
Soudry D and Carmon Y 2016 {\em arXiv preprint arXiv:1605.08361\/}

\bibitem{cooper2018loss}
Cooper Y 2018 {\em arXiv preprint arXiv:1804.10200\/}

\bibitem{verpoort2020archetypal}
Verpoort P~C, Wales D~J {\em et~al.\/} 2020 {\em Proceedings of the National
  Academy of Sciences\/} {\bf 117} 21857--21864

\bibitem{janocha2017loss}
Janocha K and Czarnecki W~M 2017 {\em arXiv preprint arXiv:1702.05659\/}

\bibitem{kornblith2020s}
Kornblith S, Lee H, Chen T and Norouzi M 2020 {\em arXiv preprint
  arXiv:2010.16402\/}

\bibitem{ballard2017energy}
Ballard A~J, Das R, Martiniani S, Mehta D, Sagun L, Stevenson J~D and Wales D~J
  2017 {\em Physical Chemistry Chemical Physics\/} {\bf 19} 12585--12603

\bibitem{mannelli2019afraid}
Mannelli S~S, Biroli G, Cammarota C, Krzakala F and Zdeborov{\'a} L 2019 {\em
  arXiv preprint arXiv:1907.08226\/}

\bibitem{arous2019landscape}
Arous G~B, Mei S, Montanari A and Nica M 2019 {\em Communications on Pure and
  Applied Mathematics\/} {\bf 72} 2282--2330

\bibitem{goldblum2019truth}
Goldblum M, Geiping J, Schwarzschild A, Moeller M and Goldstein T 2019 {\em
  arXiv preprint arXiv:1910.00359\/}

\bibitem{niroomand2024explainable}
Niroomand M~P, Dicks L, Pyzer-Knapp E and Wales D~J 2024 {\em Machine Learning:
  Science and Technology\/}

\bibitem{niroomand2024insights}
Niroomand M~P, Dicks L, Pyzer-Knapp E~O and Wales D~J 2024 {\em Digital
  Discovery\/} {\bf 3} 637--648

\bibitem{franz2016simplest}
Franz S and Parisi G 2016 {\em Journal of Physics A: Mathematical and
  Theoretical\/} {\bf 49} 145001

\bibitem{geiger2019jamming}
Geiger M, Spigler S, d'Ascoli S, Sagun L, Baity-Jesi M, Biroli G and Wyart M
  2019 {\em Physical Review E\/} {\bf 100} 012115

\bibitem{franz2019jamming}
Franz S, Hwang S and Urbani P 2019 {\em Physical review letters\/} {\bf 123}
  160602

\bibitem{franz2019critical}
Franz S, Sclocchi A and Urbani P 2019 {\em Physical review letters\/} {\bf 123}
  115702

\bibitem{geiger2020scaling}
Geiger M, Jacot A, Spigler S, Gabriel F, Sagun L, d’Ascoli S, Biroli G,
  Hongler C and Wyart M 2020 {\em Journal of Statistical Mechanics: Theory and
  Experiment\/} {\bf 2020} 023401

\bibitem{geiger2020perspective}
Geiger M, Petrini L and Wyart M 2020 {\em arXiv preprint arXiv:2012.15110\/}

\bibitem{ruiz2017bifurcation}
Ruiz-Garcia M, Bonilla L and Prados A 2017 {\em Physical Review E\/} {\bf 96}
  062147

\bibitem{rocks2017designing}
Rocks J~W, Pashine N, Bischofberger I, Goodrich C~P, Liu A~J and Nagel S~R 2017
  {\em Proceedings of the National Academy of Sciences\/} {\bf 114} 2520--2525

\bibitem{ruiz2019tuning}
Ruiz-Garc{\'\i}a M, Liu A~J and Katifori E 2019 {\em Physical Review E\/} {\bf
  100} 052608

\bibitem{yan2017architecture}
Yan L, Ravasio R, Brito C and Wyart M 2017 {\em Proceedings of the National
  Academy of Sciences\/} {\bf 114} 2526--2531

\bibitem{ruiz2021emergent}
Ruiz-Garc{\'\i}a M and Katifori E 2021 {\em Physical Review E\/} {\bf 103}
  062301

\bibitem{martinez2024fluidic}
Mart{\'\i}nez-Calvo A, Biviano M~D, Christensen A~H, Katifori E, Jensen K~H and
  Ruiz-Garc{\'\i}a M 2024 {\em Nature Communications\/} {\bf 15} 3121

\bibitem{altman2025collective}
Altman L~E, Awad N, Durian D~J, Ruiz-Garcia M and Katifori E 2025 {\em arXiv
  preprint arXiv:2502.05570\/}

\bibitem{ruiz2020topologically}
Ruiz-Garcia M and Katifori E 2020 {\em arXiv preprint arXiv:2001.01811\/}

\bibitem{sagun2017empirical}
Sagun L, Evci U, Guney V~U, Dauphin Y and Bottou L 2017 {\em arXiv preprint
  arXiv:1706.04454\/}

\bibitem{sagun2016eigenvalues}
Sagun L, Bottou L and LeCun Y 2016 {\em arXiv preprint arXiv:1611.07476\/}

\bibitem{sankar2021deeper}
Sankar A~R, Khasbage Y, Vigneswaran R and Balasubramanian V~N 2021 A deeper
  look at the hessian eigenspectrum of deep neural networks and its
  applications to regularization {\em Proceedings of the AAAI Conference on
  Artificial Intelligence\/} vol~35 pp 9481--9488

\bibitem{yao2020pyhessian}
Yao Z, Gholami A, Keutzer K and Mahoney M~W 2020 Pyhessian: Neural networks
  through the lens of the hessian {\em 2020 IEEE international conference on
  big data (Big data)\/} (IEEE) pp 581--590

\bibitem{ahn2022understanding}
Ahn K, Zhang J and Sra S 2022 Understanding the unstable convergence of
  gradient descent {\em International Conference on Machine Learning\/} (PMLR)
  pp 247--257

\bibitem{kaur2023maximum}
Kaur S, Cohen J and Lipton Z~C 2023 On the maximum hessian eigenvalue and
  generalization {\em Proceedings on\/} (PMLR) pp 51--65

\bibitem{xing2018walk}
Xing C, Arpit D, Tsirigotis C and Bengio Y 2018 {\em arXiv preprint
  arXiv:1802.08770\/}

\bibitem{cohen2022gradient}
Cohen J~M, Kaur S, Li Y, Kolter J~Z and Talwalkar A 2022 Gradient descent on
  neural networks typically occurs at the edge of stability (\textit{Preprint}
  \eprint{2103.00065})

\bibitem{zhu2022understanding}
Zhu X, Wang Z, Wang X, Zhou M and Ge R 2022 {\em arXiv preprint
  arXiv:2210.03294\/}

\bibitem{arora2022understanding}
Arora S, Li Z and Panigrahi A 2022 Understanding gradient descent on the edge
  of stability in deep learning {\em International Conference on Machine
  Learning\/} (PMLR) pp 948--1024

\bibitem{ruiz2021tilting}
Ruiz-Garcia M, Zhang G, Schoenholz S~S and Liu A~J 2021 Tilting the playing
  field: Dynamical loss functions for machine learning {\em International
  Conference on Machine Learning\/} (PMLR) pp 9157--9167

\bibitem{murugan2019bioinspired}
Murugan A and Jaeger H~M 2019 {\em MRS Bulletin\/} {\bf 44} 96--105

\bibitem{murugan2021roadmap}
Murugan A, Husain K, Rust M~J, Hepler C, Bass J, Pietsch J~M, Swain P~S, Jena
  S~G, Toettcher J~E, Chakraborty A~K {\em et~al.\/} 2021 {\em Physical
  biology\/} {\bf 18} 041502

\bibitem{mustonen2009fitness}
Mustonen V and L{\"a}ssig M 2009 {\em Trends in genetics\/} {\bf 25} 111--119

\bibitem{di2024emergent}
Di~Bari L, Bisardi M, Cotogno S, Weigt M and Zamponi F 2024 {\em bioRxiv\/}
  2024--03

\bibitem{cairns2022strong}
Cairns J, Borse F, Mononen T, Hiltunen T and Mustonen V 2022 {\em Evolution
  Letters\/} {\bf 6} 266--279

\bibitem{falk2023learning}
Falk M~J, Wu J, Matthews A, Sachdeva V, Pashine N, Gardel M~L, Nagel S~R and
  Murugan A 2023 {\em Proceedings of the National Academy of Sciences\/} {\bf
  120} e2219558120

\bibitem{bengio2009curriculum}
Bengio Y, Louradour J, Collobert R and Weston J 2009 Curriculum learning {\em
  Proceedings of the 26th annual international conference on machine
  learning\/} pp 41--48

\bibitem{pmlr-v48-amodei16}
Amodei D, Ananthanarayanan S, Anubhai R, Bai J, Battenberg E, Case C, Casper J,
  Catanzaro B, Cheng Q, Chen G, Chen J, Chen J, Chen Z, Chrzanowski M, Coates
  A, Diamos G, Ding K, Du N, Elsen E, Engel J, Fang W, Fan L, Fougner C, Gao L,
  Gong C, Hannun A, Han T, Johannes L, Jiang B, Ju C, Jun B, LeGresley P, Lin
  L, Liu J, Liu Y, Li W, Li X, Ma D, Narang S, Ng A, Ozair S, Peng Y, Prenger
  R, Qian S, Quan Z, Raiman J, Rao V, Satheesh S, Seetapun D, Sengupta S,
  Srinet K, Sriram A, Tang H, Tang L, Wang C, Wang J, Wang K, Wang Y, Wang Z,
  Wang Z, Wu S, Wei L, Xiao B, Xie W, Xie Y, Yogatama D, Yuan B, Zhan J and Zhu
  Z 2016 Deep speech 2 : End-to-end speech recognition in english and mandarin
  {\em Proceedings of The 33rd International Conference on Machine Learning\/}
  ({\em Proceedings of Machine Learning Research\/} vol~48) ed Balcan M~F and
  Weinberger K~Q (New York, New York, USA: PMLR) pp 173--182

\bibitem{graves2016hybrid}
Graves A, Wayne G, Reynolds M, Harley T, Danihelka I, Grabska-Barwi{\'n}ska A,
  Colmenarejo S~G, Grefenstette E, Ramalho T, Agapiou J {\em et~al.\/} 2016
  {\em Nature\/} {\bf 538} 471--476

\bibitem{silver2017mastering}
Silver D, Schrittwieser J, Simonyan K, Antonoglou I, Huang A, Guez A, Hubert T,
  Baker L, Lai M, Bolton A {\em et~al.\/} 2017 {\em nature\/} {\bf 550}
  354--359

\bibitem{lozano2025towards}
Lozano S, Spitzer M, Strittmatter Y, Moeller K and Ruiz-Garcia M 2025 Towards a
  curriculum for neural networks to simulate symbolic arithmetic {\em
  Proceedings of the Annual Meeting of the Cognitive Science Society\/} vol~47

\bibitem{strittmatter2025curriculum}
Strittmatter Y, Sarao~Mannelli S, Ruiz-Garcia M, Musslick S and Spitzer M 2025
  Curriculum learning in humans and neural networks {\em Proceedings of the
  Annual Meeting of the Cognitive Science Society\/} vol~47

\bibitem{belkin2019reconciling}
Belkin M, Hsu D, Ma S and Mandal S 2019 {\em Proceedings of the National
  Academy of Sciences\/} {\bf 116} 15849--15854

\bibitem{zhang2021understanding}
Zhang C, Bengio S, Hardt M, Recht B and Vinyals O 2021 {\em Communications of
  the ACM\/} {\bf 64} 107--115

\bibitem{nakkiran2021deep}
Nakkiran P, Kaplun G, Bansal Y, Yang T, Barak B and Sutskever I 2021 {\em
  Journal of Statistical Mechanics: Theory and Experiment\/} {\bf 2021} 124003

\bibitem{d2020double}
d’Ascoli S, Refinetti M, Biroli G and Krzakala F 2020 Double trouble in
  double descent: Bias and variance (s) in the lazy regime {\em International
  Conference on Machine Learning\/} (PMLR) pp 2280--2290

\bibitem{rocks2022memorizing}
Rocks J~W and Mehta P 2022 {\em Physical review research\/} {\bf 4} 013201

\bibitem{gur2018gradient}
Gur-Ari G, Roberts D~A and Dyer E 2018 {\em arXiv preprint arXiv:1812.04754\/}

\bibitem{annesi2023star}
Annesi B~L, Lauditi C, Lucibello C, Malatesta E~M, Perugini G, Pittorino F and
  Saglietti L 2023 {\em Physical Review Letters\/} {\bf 131} 227301

\bibitem{lewkowycz2020large}
Lewkowycz A, Bahri Y, Dyer E, Sohl-Dickstein J and Gur-Ari G 2020 {\em arXiv
  preprint arXiv:2003.02218\/}

\bibitem{mccloskey1989catastrophic}
McCloskey M and Cohen N~J 1989 Catastrophic interference in connectionist
  networks: The sequential learning problem {\em Psychology of learning and
  motivation\/} vol~24 (Elsevier) pp 109--165

\bibitem{ruiz2022model}
Ruiz-Garcia M 2022 {\em Scientific Reports\/} {\bf 12} 10736

\end{thebibliography}

\end{document}